\documentclass[11pt]{article}
\usepackage{acl2012}
\usepackage{times}
\usepackage{latexsym}
\usepackage{graphicx}
\usepackage{amsmath}
\usepackage{multirow}
\usepackage{url}
\usepackage{algorithm}
\usepackage{algorithmic}

\title{Supervised Fine Tuning for Word Embedding with Integrated Knowledge}

\author{Xuefeng Yang \\
  Nanyang Technological University \\
  50 Nanyang Avenue Singapore 639798 \\
  {\tt yang0302@e.ntu.edu.sg} \\\And
  Kezhi Mao \\
  Nanyang Technological University \\
  50 Nanyang Avenue Singapore 639798 \\
  {\tt ekzmao@ntu.edu.sg} \\}
  
\date{}

\begin{document}
\maketitle
\begin{abstract}
  Learning vector representation for words is an important research field which may benefit many natural language processing tasks. Two limitations exist in nearly all available models, which are the bias caused by the context definition and the lack of knowledge utilization. They are difficult to tackle because these algorithms are essentially unsupervised learning approaches. Inspired by deep learning, the authors propose a supervised framework for learning vector representation of words to provide additional supervised fine tuning after unsupervised learning. The framework is knowledge rich approacher and compatible with any numerical vectors word representation. The authors perform both intrinsic evaluation like attributional and relational similarity prediction and extrinsic evaluations like the sentence completion and sentiment analysis. Experiments results on 6 embeddings and 4 tasks with 10 datasets show that the proposed fine tuning framework may significantly improve the quality of the vector representation of words.
\end{abstract}

Learning a numerical vector to represent the semantic meaning of a word is a research topic of wide interests in computational linguistics. Various applications like information retrieval \cite{manning2008introduction}, sentiment analysis \cite{maas2011learning} and semantic role labeling \cite{collobert:2011b} benefit from the vector representation of words. There are two classes of methods for vector representation learning of words, including \emph{distributional semantic models} (DSMs) and \emph{neural language models} (word embedding). In DSMs, large sparse global co-occurrence matrix representing the context of words is first constructed , dimension reduction techniques such as SVD are then applied to find the low dimension representation of words \cite{clark2012vector,turney2010frequency}. Neural language models learn the vector representation of words using artificial neural networks. This method is firstly studied in \cite{bengio2003neural}, and many variants have been proposed \cite{turian2010word,collobert2011natural,mikolov2013efficient,mikolov2013distributed,Pennington2014glove}. In neural language methods, the key issue is the formulation of the training target function, minimization or maximization of which may produce meaningful vector representation of words. Ideally, the training target function should reflect the objective of word representation learning, that is the semantic similarity between words represented by distance measures between word vectors is consistent with human cognition. The training target in the above works is to maximise the context prediction ability which is not directly related to the word representations learning objective, therefore they are essentially unsupervised approaches working with a pseudo supervised trick.

In both DSMs and neural language models, human defined context for each word is needed. The selection of context may have a strong influence on the word vectors obtained, however, there is no generally best context definition because emphasizing one perspective of context will result in the lack of other valuable information. The examples in Table \ref{context} show that the context length have a impact on the learned word vectors. The three models are trained with the same settings except the size of context window.

\begin{table}[!htb]
\centering
\caption{Top 5 nearest neighbours for ``snake".}
\label{context}

\begin{tabular}{|c c c|}
\hline
window size 5 & window size 7 & window size 9 \\
\hline
tarantula & cobra & cobra\\
venomous & tarantula & lizard\\
frog & nonvenomous &  tarantula \\
rattlesnake & spider & frog \\
lizard & python & porcupine \\
\hline
\end{tabular}
\end{table}

Many research works show that the marriage of unsupervised and supervised learning may achieve better performance. For example, in deep learning, unsupervised feature learning are employed to initialize the neural network to enhance the performance. This inspires the authors to introduce supervised fine tuning framework to the word representation learning, with the goal of addressing the bias problem resulted from context definition.

However, introducing supervised learning into word representation learning is not a trivial issue. The most challenging problem is how to automatically generate labeled data. Manually labelling data is impractical because it is difficult to determine the exact numerical value for semantic similarity, and the size of required training data is very large. Although lexical semantic resources are available, the knowledge graph is not directly applicable for supervised learning. Another challenging problem is how to design a suitable learning algorithm so that the fine tuning will not disrupt too much of the context-based learning results.

The supervised fine tuning learning framework proposed in this paper provides solutions to the above mentioned problems. The core idea behind the proposed framework is to use the ranking of word similarities instead of the exact similarity measure values as the training target. First, an algorithm that automatically generates labeled data based on existing word embeddings and knowledge resources is proposed. Second, an inverse error weighted mini-batch stochastic gradient descent optimization algorithm is designed, which can effectively absorb the complementary knowledge to fine tune the word embeddings without disrupting the original geometric similarity of word embeddings.

The remainder of the paper is organized as follows. In Section $2$, related work on word representation learning is briefly reviewed. The proposed supervised fine tuning framework is detailed in Section $3$. Evaluation and model analysis are given in Section $4$. Section $5$ concludes the paper.

\section{Related Work}
Vector representation learning for words has received considerable attentions in the past two decades. The methods reported in the literature can be classified into two categories, including distributional semantic models and neural language models.

Distributional semantic models are based on the principle that words with similar semantic meanings have similar contextual information \cite{harris1954distributional,firth57synopsis}. The statistical contextual information is usually summarized into a large sparse matrix, each row of which is the global contextual information of a word in the large corpus, where weighting scheme like tf-idf and PMI are often used to remove the bias of frequency. Dimension reduction techniques like singular value decomposition is then applied to the high dimensional sparse matrix to generate a low dimensional dense matrix with the same number of rows as the original sparse matrix. Each row of the obtained low dimensional matrix is the vector representation of a word.

The neural language models are based on artificial neural networks, whose input is the context of a word and the parameters are the vector representation of the words. The parameters of the neural network are adjusted based on the training target and approximation algorithm. A variety of training targets have been proposed. For example, \cite{bengio2003neural} employs sequence of words to predict the next word, \cite{turian2010word} employs context to predict the middle word, and \cite{mikolov2013efficient} predicts all context given the central word. For the approximation algorithm, hierarchical softmax is borrowed from \cite{Morin05hierarchicalprobabilistic} and different methods to construct the hierarchical tree are studied in \cite{mnih2009scalable,mikolov2013distributed}, negative sampling series approaches are proposed in \cite{collobert2011natural,Mnih12afast,mikolov2013distributed}.

Besides the above data-based neural language models, there are methods that incorporates knowledge into the learning of neural language models. The major training target of these knowledge powered models is still based on the contextual information, and the knowledge is used as auxiliary resources. For example, \cite{Xu:2014:RGF:2661829.2662038} propose a framework to add relational and categorical knowledge as regularization of the original training target, and \cite{C14-1015} utilize the morphological knowledge as both additional input representation and auxiliary supervision. \cite{Botha2014} trained a compositional morphological vector representation, in which a word vector is the addition of its morphological factor vectors. In spite of the introduction of knowledge into the learning algorithm, these methods still suffer from the bias problem caused by the context definition. \cite{DBLP:conf/conll/PassosKM14} proposed a lexicon infused word representation learning algorithm for named entity recognition which is built upon the existing skip-gram model.

\cite{bansal2014tailoring} employ an ensemble on different word representations which outperforms all individual word representation on the dependency parsing application, this suggests that the complementary information exists in different word representations. However, employing ensemble classifier may dramatically increase the computational burden in prediction because it utilize multiple classifiers. The supervised fine tuning framework proposed in this paper may be treated as a compressed ensemble model, which directly compresses the complementary information into an individual word representation. Instead of training ensemble classifier on all the word embeddings, we may use the complementary information by training an individual classifier on the new word embedding.

\section{Supervised Fine Tuning for Word Representations}

\subsection{Overview}
\begin{figure}[!htb]
\begin{center}
\centerline{\includegraphics[width=\columnwidth]{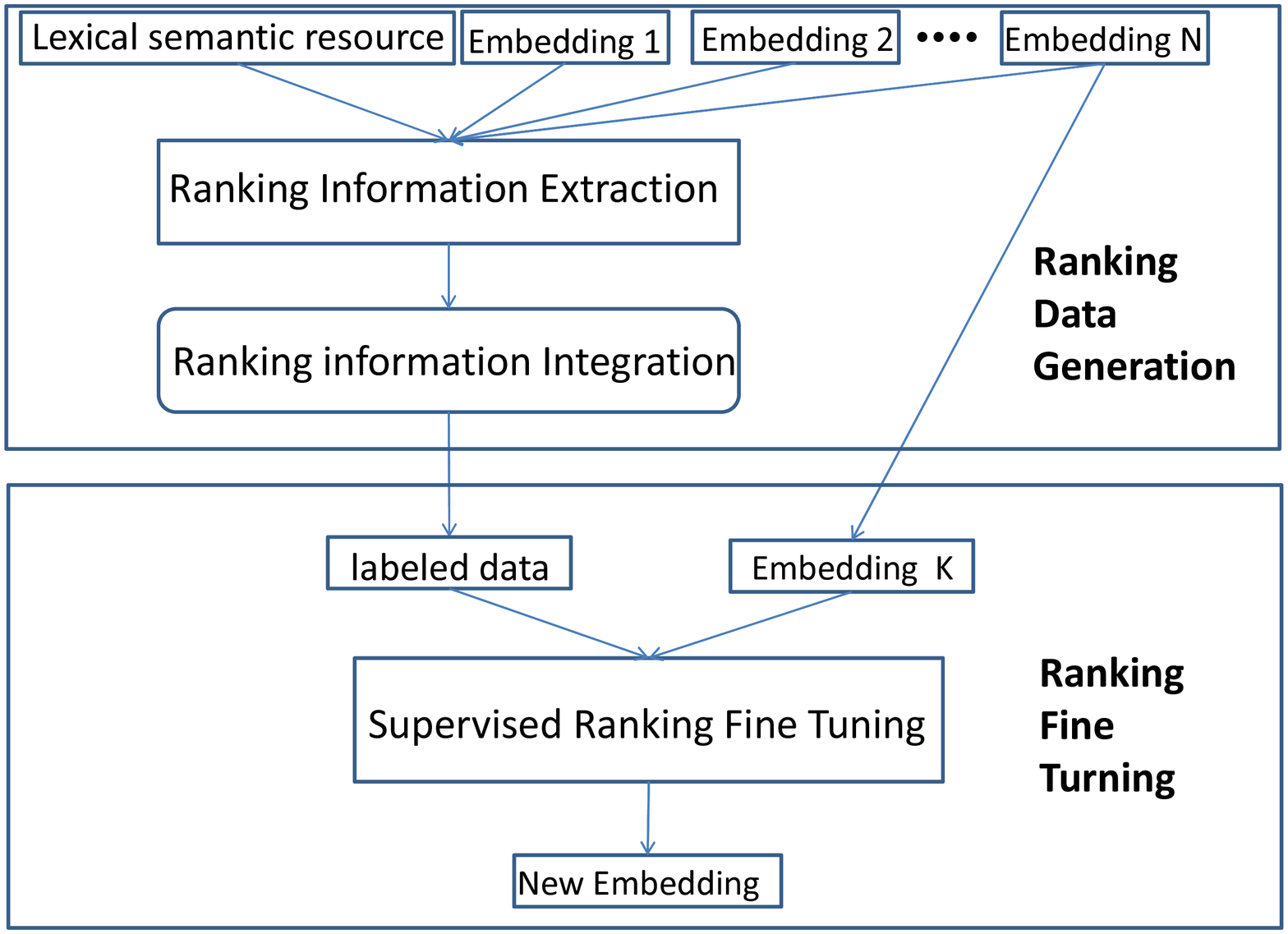}}
\caption{Supervised Fine Tuning Framework.}
\label{overview}
\end{center}
\end{figure}

To alleviate the bias problem mentioned above, this study propose a supervised fine tuning framework for word representation learning. The overview of the framework is shown in Figure \ref{overview}. The proposed supervised fine tuning framework is build upon the established word embedding and lexical semantic resources. There are two parts in the fine tuning framework, including ranking data generation and supervised ranking fine tuning.

In the ranking data generation phrase, the ranking information for every word is extracted from each individual word embedding, which is then integrated by a score based multiple criteria fusion algorithm. Human summarized knowledge like WordNet \cite{miller1995wordnet} may also be included in the data generation phrase. With the labeled training data obtained above, the supervised ranking learning algorithm may fine tune the original word vectors to encode the complementary knowledge from other word embeddings.

\subsection{Automatic Labeling of Training Data}
As mentioned in introduction, the training target is to learn the semantic similarity ranking instead of similarity measure itself. The goal of training data labeling is to generate the ranking of semantically similar words for each training word.

\subsubsection{Why Ranking ?}

The biggest challenge for supervised word representation fine tuning is the labelling of training data. First, it is difficult to define the exact similarity values between two words. Second, it is impractical to manually label a huge number of training data to support supervised word representation learning. In the previous studies, each method starts from scratch, without using the results of other embeddings. In contrast, our work is built upon existing word embeddings. Thus, labelling of training data can be conducted automatically using the existing word embeddings.

The similarity measure is affected by many factors such as the dimensionality of the word vectors, the employed learning algorithms and the corpus size. An example from GloVe embedding \cite{Pennington2014glove} may illustrate the phenomenon. The cosine similarity value between words ``fish" and ``salmon" in Glove embedding 300 dimension version is 0.6596, and the value in Glove embedding 50 dimension version is 0.8340. Although the similarity values are quite different, the word ``salmon" is the most similar word to word ``fish" in both embeddings. This reveals that the ranking of similarity values is more robust than the similarity values itself. Inspired by this finding, the authors propose to employ ranking information as the supervised training targets

\subsubsection{Multiple Ranking Integration}

The similarity ranking information can be extracted from existing word embeddings and is used as the training target. But the issue is which word embedding we should use. Even the state-of-art embeddings may not always provide reliable ranking information. Table \ref{noise} shows some obvious errors in the related words of words ``hill", ``run", and ``paper" in the GloVe word embedding. The words in bold are the errors between the real related words and the number beside the word represents the ranking position. Obviously, errors exist even in the highly frequent words like ``paper" and ``hill". To address this problem, the authors propose to enhance ranking information label using multiple word embeddings. The employed multiple word embeddings should be trained with different algorithms, context definition and corpus. The complementary effect of different word embeddings will improve robustness and reliability of the ranking information.

\begin{table}[!htb]
\centering
\caption{Errors in GloVe300 embedding.}
\label{noise}
\begin{tabular}{c c c}
Hill & Run & Paper\\
\hline
\textbf{now}, \textbf{57} & \textbf{three}, \textbf{12} & \textbf{instead}, \textbf{30} \\
\textbf{known}, \textbf{64} & \textbf{only}, \textbf{17} & \textbf{put}, \textbf{33} \\
woods, 70 & \textbf{not}, \textbf{30} & document,35 \\
hillside, 72 & come, 39 & books, 42 \\
\textbf{called}, \textbf{74} & \textbf{seven}, \textbf{47} & \textbf{mirror}, \textbf{53}\\
wood, 97 & take, 58 & publish, 57 \\
\end{tabular}
\end{table}

To integrate multiple ranking information sources, a score-based multiple criteria fusion algorithm is utilized. Generally, in score-based multiple criteria fusion algorithm, a common used score is defined to measure the credit of each candidate data in every criteria. A combination algorithm is then utilized to aggregate the multiple score into one consensus score. Finally a ranking procedure is performed to rank the candidate data based on their consensus scores. In this study, the candidate data is the candidate words to be selected as labeled data, and the score to measure the credit of the candidate words is a normalized cosine similarity measure.

\begin{equation}
\setlength{\abovedisplayskip}{2.5pt}
\setlength{\belowdisplayskip}{2.5pt}
 credit^{\omega_{r}}(\omega_{d}) = \frac{cosine(\omega_{r},\omega_{d})}{\max\{cosine(\omega_{r},\omega)|\omega \in V\}}
\label{credit}
\end{equation}

The score of word $\omega_{d}$ in the ranking for word $\omega_{r}$ may be obtained by Equation (\ref{credit}), where $V$ represents the whole vocabulary. In Equation (\ref{dividetime}), the cumulated score for each candidate word in all embeddings is divided by the times the word pair appearing in these embeddings.

\begin{equation}
\setlength{\abovedisplayskip}{2.5pt}
\setlength{\belowdisplayskip}{2.5pt}
 score^{\omega_{r}}(\omega_{d}) = \frac{\sum_{E\in Embeds} credit_{E}^{\omega_{r}}(\omega_{d})}{\sum_{E\in Embeds} I((\omega_{r}, \omega_{d}) \in E)}
\label{dividetime}
\end{equation}

Where the $I$ represents the indicator function, whose value is 1 if the input condition is satisfied. The data with high averaged score is believed to be the trustable knowledge, while the low score data should be the bias or errors.

\subsubsection{Algorithm}
The whole procedure of the labeled data generation and integration is given in Algorithm \ref{alg2}\footnote{The authors' implementation is based on sparse matrix and nested mapping because the size of matrix is very large.}.

\begin{algorithm}[!htb]
   \caption{Labeled Data Generation and  Integration}
   \label{alg2}
\begin{algorithmic}
   \STATE \textbf{Input}: VocabList $V$ stores words index in a list
   \STATE 2-D Count Matrix $D$ corresponding to the word in $V$
   \STATE 2-D Matrix $C$ stores the times of word pair selected
   \STATE 2-D Matrix $S$ stores the times of word pair appearing
   \STATE List of normalized embeddings: $embeddings$
   \STATE Number of words to extract $N$
   \STATE $data$ is a dictionary data structure stores the output
   \STATE \textbf{Initialize} $D = 0$, $C = 0$
   \STATE \textbf{1 Calculate the score of data}
   \FORALL{ $em$ in $embeddings$}
   \FORALL{ $w_{v}$ in $V$}
   \STATE $(w_{N},s_{N})$ $\gets$ top $N$ of $\{cos_{em}(w_{v},w_{i})|i \in V\}$
   \STATE $s_{max}$ $\gets$ $max(s_{N})$
   \FORALL{$w_{i}$ and $s_{i}$ in $w_{N},s_{N}$}
   \STATE $D[w_{v},w_{i}] += s_{i}/s_{max}$
   \STATE $C[w_{v},w_{i}] += 1$
   \ENDFOR
   \ENDFOR
   \ENDFOR
   \STATE \textbf{2 Remove low score data}
   \FORALL{$w_{v1}$ in $V$}
   \FORALL{$w_{v2}$ in $V$}
   \IF{$C[w_{v1},w_{v2}] <= 2$}
   \STATE $D[w_{v1},w_{v2}] = 0$
   \ENDIF
   \ENDFOR
   \ENDFOR
   \STATE \textbf{3 Remove bias of frequency}
   \STATE $D = D / S$     hint: point wise division
   \FORALL{$w_{v}$ in $V$}
   \STATE $data[w_{v}]$ $\gets$ select nonzeros in $D[w_{v}]$ and sort
   \ENDFOR
   \STATE \textbf{Output}: $data$ stores the ranking of related words
\end{algorithmic}
\end{algorithm}

\subsubsection{Reduction of Unnecessary Data}
It is impractical to learn the full ranking for each word because the whole similarity matrix is extremely large. The authors find that it is not necessary to employ full ranking as training data for each word. Intuitively, a word is unrelated to most of the other words in the vocabulary, the word embeddings should also be consistent with this fact that most of the similarity values should indicate two words are unrelated. Additionally, adjusting whether ``pen" or ``water" is more similar to ``basketball" does not make any sense since they are all totally unrelated. Based on the analysis above, it is reasonable to only care the meaningful word pairs with strong relations between each other and ignore other unrelated words. Therefore, only the top ranked words in the ranking deserve further fine tuning. Without the consideration of the unrelated words, the data size of the supervised training may be greatly reduced to hundreds per words. In this study, the authors extract the top 200 most similar words as the candidate data for each word from each individual word embedding based on the empirical experience.

\subsection{Supervised Ranking Fine Tuning}

\subsubsection{Training Target}
As mentioned in the Introduction, the training target in most neural language models is to maximize the context prediction ability of word embeddings. Different from the previous approachers, the proposed fine tuning framework attempts to adjust the word vectors of a particular word embeddings so that the word similarity ranking is in line with the labeled data. Fitting the ranking of similarity measure may directly alleviate the pain caused by the bias of context, therefore it is a supervised solution to the bias problem caused by context definition. In addition, the ranking of semantic similarity is the most important property of the desired word representation in most tasks. In this sense, it is a supervised solution to the fine tuning of word embeddings.

To achieve the above goal, a ranking loss function is employed as the cost function. The ranking loss function $J_{rank}$ is shown in Equation (\ref{loss}).

\begin{equation}
\setlength{\abovedisplayskip}{2.5pt}
\setlength{\belowdisplayskip}{2.5pt}
J_{rank} = \sum_{\omega_{v} \in V}\sum_{\omega_{r} \in D_{\omega_{v}}}\mid L_{\omega_{r}} - R_{\omega_{r}}\mid^{2}
\label{loss}
\end{equation}

Where $V$ is the vocabulary and $D_{\omega_{v}}$ is the related data set of word $\omega_{v}$, $L$ and $R$ denote the ranking in the labeled data and the word embedding under study, respectively.

Because the ranking loss is not differentiable, the authors choose to minimize the semantic similarity loss between the desired ranking position and the real ranking position. Given the desired ranking position, the similarity value corresponding to the desired ranking position is employed as the real training target. Minimizing the difference of similarity values between desired position and real position may also reduce the ranking loss. The similarity value loss function $J_{simi}$ is given in Equation \ref{similoss}, where $S_{\omega_{v}}$ denotes the sorted similarity values for word $\omega_{v}$

\begin{equation}
\setlength{\abovedisplayskip}{2.5pt}
\setlength{\belowdisplayskip}{2.5pt}
J_{simi}\!=\!\sum_{\omega_{v} \in V}\!\sum_{\omega_{r} \in D_{\omega_{v}}}\!\mid
\!S_{\omega_{v}}(L_{\omega_{r}})-S_{\omega_{v}}(R_{\omega_{r}})\mid^{2}
\label{similoss}
\end{equation}

\subsubsection{Inverse error Weighted Mini-batch SGD Optimization}

Stochastic gradient descent(SGD) is a widely used optimization algorithm to minimize the loss function. The iterative update rule for parameters is shown in Equation (\ref{sgd}).

\begin{equation}
\setlength{\abovedisplayskip}{2.5pt}
\setlength{\belowdisplayskip}{2.5pt}
\omega = \omega - \eta \times\frac{\partial J(d_{i})}{\partial \omega}
\label{sgd}
\end{equation}

Where $\omega$ and $\eta$ denote the parameters and learning rate, and $\frac{\partial J(d_{i})}{\partial \omega}$ is the approximation of gradient of the loss function $J$ based on data $d_{i}$. Because the computation of similarity distribution is very expensive, mini-batch SGD is employed to speed up the learning process. In mini-batch SGD, the gradient of the loss function is approximated by a batch of data, which is shown in Equation (\ref{minibatch}).

\begin{equation}
\setlength{\abovedisplayskip}{2.5pt}
\setlength{\belowdisplayskip}{2.5pt}
\begin{split}
g_{batch} &= \frac{1}{N}\sum_{i \in batch} \frac{\partial J(d_{i})}{\partial \omega} \\
\omega &= \omega - \eta \times g_{batch}
\end{split}
\label{minibatch}
\end{equation}

Where $N$ is the size of batch and $g_{batch}$ denotes the batch gradient. However, the direct applying mini-batch SGD encounters a serious problem: the training loss is reduced, but the performance on all tasks deteriorates. This is quite similar to the overfitting problem in which the models learn the noise of the training data and perform badly on unknown testing dateset. However, there are two differences between this problem and overfitting. The overfitting phenomenon occurs at the last phase and the model generally should have more parameters than necessary, this generalization decreasing problem happens at the start of the training and it exists even the size of parameters is very small.

After exploration of the learning process, the authors find that the reason behind this problem is that the external knowledge of ranking information may disrupt the geometric similarity of the original word embedding under the standard mini-batch SGD optimization algorithm. The existing word embeddings are sill far from perfect, and some strongly related words in reality may be ignored, which may be caused by many factors such as polysemy, corpus bias and imbalanced word frequency. Such words may have very large ranking errors, then the gradient of these words become very large because the magnitude of the gradient is proportional to the error. Finally, the gradient of these words dominates the overall gradient and disrupt the original learned geometric similarity.

In this fine tuning framework, the principle that the data with larger error deserves more learning is not applicable, the authors propose an inverse error weighted mini-batch SGD optimization algorithm to remove the effect of the large errors.

\begin{equation}
\setlength{\abovedisplayskip}{2.5pt}
\setlength{\belowdisplayskip}{2.5pt}
g_{batch} = \frac{1}{N}\sum_{i \in batch} \frac{1}{e(d_{i})}  \frac{\partial J(d_{i})}{\partial \omega}
\label{errorfree}
\end{equation}

Equation \ref{errorfree} shows the inverse error weighted gradient, where $e(d_{i})$ denotes the error of data $d_{i}$.

Besides the update from the labeled data, the algorithm also deals with the data that are in the meaningful range but not covered by the labeled data. These data are close to the trained word in the word embedding, but actually not related to the word in reality. These data are also processed by the proposed inverse error weighted mini-batch SGD, but the errors are not included in the cost function.

To determine the range of meaningful semantic similarity in original word embeddings, a threshold based on random similarity distribution is utilized. The idea behind this is that the similarity values which are difficult to be randomly generated are the real knowledge the word embedding learned. To calculate the threshold for specific word embedding, a random matrix following uniformly distribution and having the same shape with the word embedding is generated, then the mean of all the similarity values in the top 5 ranking for all words is calculated as the random threshold. To speed up the generation process, a sample may be used to approximate the whole distribution. The similarity values above the random threshold have a large probability to be meaningful data.

The detail of the updating rules is provided in Algorithm \ref{alg3}.

\begin{algorithm}[!htb]
   \caption{Ranking Learning Algorithm}
   \label{alg3}
\begin{algorithmic}
   \STATE \textbf{Input}: vocabList $V$ stores words index in a list and $em$ is the initial embedding
   \STATE $data$ is a two tier nested mapping stores all training data
   \STATE $data_{w_{v}}$ and $rank_{w_{v}}$ are mappings between words and their ranking for $w_{v}$ in supervised label and trained embedding respectively
   \STATE $ul$ is a list stores all the local update for word $w_{v}$
   \STATE $update$ is the global update vector for word $w_{v}$
   \STATE $\delta$ is the random threshold and $\sigma$ is the learning rate
   \STATE \textbf{Initialize} $error = len(V) \times d$, $stop = False$
   \WHILE {$stop$ != True}
   \FORALL {$w_{v}$ in $V$}
   \STATE $data_{w_{v}}$ $\gets$ $data[w_{v}]$
   \STATE $rank_{w_{v}}$ $\gets$ get ranking of all words for $w_{v}$
   \STATE $ul$ $\gets$ ${\o}$
   \FORALL {$W_{d}$ in $data_{w_{v}}$}
   \STATE $sign$ $\gets$ \emph{I}($rank_{w_{v}}[w_{d}] - data_{w_{v}}[w_{d}]$)
   \STATE $ul$.add($sign$ $\times$ $\cos$($w_{v},w_{d}$) $\times$ $em[w_{d}]$)
   \ENDFOR
   \STATE $ne_{w_{v}}$ $\gets$ select $rank_{w_{v}}$ if cosine$>$$\delta$ and $not$ $in$ $data_{w_{v}}$
   \FORALL {$W_{d}$ in $ne_{w_{v}}$}
   \STATE $ul$.add($-1$ $\times$ $\cos$($w_{v},w_{d}$) $\times$ $em[w_{d}]$)
   \ENDFOR
   \STATE $update$ $\gets$ mean of vectors in $ul$
   \STATE $update$ $\gets$ normalization($update$)
   \STATE $em[w_{v}]$ = $em[w_{v}]$ + $\sigma \times update$
   \STATE $em[w_{v}]$ = normalization($em[w_{v}]$)
   \ENDFOR
   \ENDWHILE
\end{algorithmic}
\end{algorithm}
\subsection{Stopping Criterion and Overfitting}
Overfitting is a frequently encountered problem in machine learning. Generally, overfitting is related to the number of parameters. If the size of parameters is larger than the potential patterns in the observations, the model may suffer from the overfitting problem. The overfitting problem is more serious in this study because the automatically generated labeled data may contain some noise and confliction. The widely employed stopping criterion is given in Equation (\ref{stop1}).

\begin{equation}
\setlength{\abovedisplayskip}{2.5pt}
\setlength{\belowdisplayskip}{2.5pt}
\frac{J_{rank}(i) - J_{rank}(i+1)}{J_{rank}(i)} <= \epsilon
\label{stop1}
\end{equation}

Where $J_{rank}(i)$ is the overall ranking loss in the $i$th epoch, which is the same as $J_{rank}$ in Equation (\ref{loss}), and$\epsilon$ is a very small value like 0.01. If the relative improvement is smaller than the predefined value $\epsilon$, the training should stop. However, the authors find that the criterion may not generalize well for the involved embeddings and tasks, even $\epsilon$ is changed according to the number of parameters. The authors employ an alternative stopping criterion, which is related to the initial performance of the word embedding. Equation (\ref{stop2}) shows the criterion, and the $J_{simi}(0)$ is related to the performance of the embeddings on various tasks. With this criterion, the $\epsilon$ selected according to the dimensionality of word embeddings and the initial performance may work better. This may be explained by the fact that the quality of different embeddings are quite different, the poorly performed models may have more space to grow and the models with good initial status are easier to fall into overfitting.

\begin{equation}
\setlength{\abovedisplayskip}{2.5pt}
\setlength{\belowdisplayskip}{2.5pt}
\frac{J_{rank}(i) - J_{rank}(i+1)}{J_{rank}(0)} <= \epsilon
\label{stop2}
\end{equation}

\section{Experiment}
To evaluate the proposed supervised fine tuning framework, three groups of experiments are designed.

The intrinsic and extrinsic evaluations experiments are conducted to test the effect of fine tuning. 6 word embeddings are further trained by the proposed supervised fine tuning framework and evaluated with 4 tasks. The baselines are the original word embeddings which includes many reputable word embeddings such as SENNA, Word2vec and GloVe.

The effect of inverse error weighted mini-batch SGD optimization algorithm is studied in the second group experiments. Two comparison groups are involving, one uses the widely employed standard optimization algorithm and another utilizes the proposed inverse error weighted variant. The comparison uses 6 embeddings and 3 tasks with 3 classical datasets, except the updating rules, all the other settings are the same for two comparison groups.

The influence of data size on the model is studied in the last group experiments with two best embeddings and 8 datasets, and the hypothesis about the reduced data size is tested.

\subsection{Evaluation Methods}
\subsubsection{Semantic Similarity Prediction}

Measuring the consistency between machine predicted similarity values and human annotated similarity values is the most widely employed task to evaluate the quality of word embeddings. The datasets are usually constructed from crowdsourcing cognitive experiments

Many datasets are available, the authors select 5 of them to cover of different perspectives and data size. Wordsim353 \cite{finkelstein2001placing} and RG65 \cite{Rubenstein:1965:CCS:365628.365657} are the two most used dataset for semantic similarity evaluation. MEN3000 \cite{bruni2014multimodal} and Mturk771 \cite{Halawi:2012:LLW:2339530.2339751} are two recently generated large datasets. YP130 dataset \cite{Yang06verbsimilarity} is designed to evaluate semantic similarity between verbs. As in other works, spearman correlation is used to measure the consistency between human annotation and machine prediction.

\subsubsection{Analogical Reasoning}

The analogical reasoning task is designed to test the ability of models for relational similarity tasks. The Google analogical reasoning dataset is introduced by \cite{mikolov2013efficient}. The question is in the following format, Man : Women = King : ?, the answer should be the exact word ``Queen". The Microsoft analogical reasoning data set \cite{export:189726} is also utilized in this study. The major difference between these two datesets is the different types of relations they focus on.

As in other works, the nearest neighbour of the vector(Women + King - Man) exclusive of the three words in questions is selected as the candidate answer. If the candidate word is the same as the answer, the question is correctly solved. In this study, the overall accuracy is employed to evaluate the word embedding. Since the size of the learned word embedding is not as large as the one used in original study \cite{mikolov2013efficient}, the question with unknown word is not included in the experiment.

\subsubsection{Sentence Completion}

The sentence completion challenge \cite{Zweig:2012:CSA:2390940.2390944} is intended to stimulate research in the area of semantic modeling. The sentence completion questions are to select words which are meaningful and coherent in the the context of a complete sentence. In each sentence, an infrequent word is chosen as the focus of the question and four alternates candidates were chosen from a list of words suggested by an N-gram language model as disturbance. Only the original word is considered as the correct answer to the question.

In this study, the author select the candidate answer based on the average similarity values between the candidates and all the words in the sentence. The word with largest average value is selected as the answer. This is the widely employed approach when this dataset is utilized to evaluate the word embeddings.

\subsubsection{Sentiment Analysis}
Two sentence level sentiment classification datasets are used in this task to do the extrinsic evaluation. The customer reviews dataset \cite{Hu:2004:MSC:1014052.1014073} contains the reviews of 5 digital products from Amazon web site, and movie review dataset \cite{Pang+Lee:05a} includes 5331 positive and 5331 negative processed sentences from rotten tomatoes movie review web site. In both datasets, the review sentence should be classified as positive or negative attitude.

In this study, the author employ the convolution neural network (CNN) as described in \cite{kim:2014:EMNLP2014} to do the sentence level sentiment classification.

\subsection{Setting and Details}
6 word embeddings are studied in our experiments, they are HLBL50 which is proposed in \cite{MnihHinton2009}, SENNA50 \cite{collobert2011natural}, RNNLM640 \cite{mikolov2011extensions}, GloVe300 \cite{Pennington2014glove}, DocAndDep2000 \cite{dad2013}, and Word2vec \cite{mikolov2013distributed}. All embeddings are available in the websites\footnote{Senna:http://ml.nec-labs.com/senna/ \newline RNNLM:http://www.fit.vutbr.cz/~imikolov/rnnlm/ \newline HLBL:http://metaoptimize.com/projects/wordreprs/ \newline Glove:http://nlp.stanford.edu/projects/glove/ \newline DocAndDep:http://www.cs.cmu.edu/~afyshe/papers/conll2013/} except the Word2vec embeddings.

The authors train the Word2vec model by the public available Word2vec toolkit using the data from the 1 billion word language modeling benchmark dataset. Hierarchical softmax approximation algorithm and skip gram structure are employed, where the window size is set to 7 and the mini count of words is set to 10. DocAndDep models are constructed from two models which are document model and dependency model, the authors utilize them separately in this study, the document models is not further trained because the initial performance has a large gap with other embeddings.

For human summarized knowledge resources, WordNet is employed in this experiment. The directly connected words and the words in the same synsets are extracted as external knowledge. In the integration process, the weights of all relations from wordnet are treated equally as 1.

The learning rate $\sigma$ is set to 0.1 in most experiments. This is for the fast convergency because the computation of cosine similarity matrix in each epoch is very expensive. To avoid overfitting, the stopping criterion is set based on the embeddings' dimensionality and original performance, the values of $\epsilon$ for different embeddings is given in Table \ref{stop}
\begin{table}[!htb]
\centering
\caption{Stopping Criterion Setting.}
\label{stop}
\begin{tabular}{c c c c}
\hline
Name & Senna50 & Skip50 & HLBL50   \\
Value & 0.04 & 0.05 & 0.004 \\
Name & Glove300 & RNNLM640 & Dep1000 \\
Value & 0.07 & 0.07 & 0.10 \\
\hline
\end{tabular}
\end{table}

The hyper-parameter setting for convolution neural network used in sentiment analysis follows the default setting utilized in \cite{kim:2014:EMNLP2014}

\subsection{Result and Analysis}
\subsubsection{Intrinsic and Extrinsic Evaluation}
\begin{table*}[!htb]
\centering
\caption{Performance of 6 word embeddings on 8 datasets.}
\label{perf}
\resizebox{\textwidth}{!}{%
\begin{tabular}{|c|c |c|c| c|c| c|c |c|c |c|c| c|}
\hline
Dataset & \multicolumn{2}{|c|}{Senna50} & \multicolumn{2}{|c|}{Skip50} & \multicolumn{2}{|c|}{HLBL50} & \multicolumn{2}{|c|}{Glove300} & \multicolumn{2}{|c|}{RNNLM640} & \multicolumn{2}{|c|}{Dep1000} \\
\hline
embedding & before & after & before & after & before & after & before & after & before & after & before & after \\
\hline
WordSim353& 0.40 &\textbf{0.56} &0.53&\textbf{0.55} &0.23&\textbf{0.53} &0.55&\textbf{0.62} &0.30&\textbf{0.55} &0.44&\textbf{0.59}\\
Mturk771 &0.48&\textbf{0.59} & 0.52&\textbf{0.60} &0.280&\textbf{0.59} &0.63&\textbf{0.69} &0.40&\textbf{0.65} &0.61&\textbf{0.70}   \\
RG65 & 0.49&\textbf{0.61} & 0.56&\textbf{0.60} &0.36&\textbf{0.66} &0.74&\textbf{0.77} & 0.51&\textbf{0.65} & 0.68&\textbf{0.76}\\
YP130 & 0.16&\textbf{0.42} &0.35&\textbf{0.45} &0.23&\textbf{0.49} &0.55&\textbf{0.59} &0.40&\textbf{0.57} & 0.56&\textbf{0.65}\\
M3k & 0.59&\textbf{0.68} &0.65&\textbf{0.73} & 0.28&\textbf{0.62} & 0.73&\textbf{0.77} & 0.46&\textbf{0.65} & 0.63&\textbf{0.73}\\
Google\_AR &0.128&\textbf{0.295} &0.202&\textbf{0.360}&0.098&\textbf{0.275} &0.479&\textbf{0.508} &0.324&\textbf{0.382} &0.206&\textbf{0.255}\\
MS\_AR & 0.185&\textbf{0.411} &0.286&\textbf{0.451} & 0.2&\textbf{0.441}  &0.665&\textbf{0.685} &0.507&\textbf{0.538} &0.393&\textbf{0.429}\\
MS\_SC & 0.329&0.289 & 0.30& 0.30 &0.26&0.254 &0.335&\textbf{0.375} &0.259&\textbf{0.291} &0.359&\textbf{0.376}\\
\hline
\end{tabular}}
\end{table*}

\begin{table}[!htb]

\caption{Performance of GloVe on Sentiment Analysis.}
\label{sa}
\resizebox{0.45\textwidth}{!}{%
\begin{tabular}{|c|c |c|c| c|c|}
\hline
Classifier & \multicolumn{2}{|c|}{static} & \multicolumn{2}{|c|}{non-static} \\
\hline
Task & Original & Trained & Original & Trained \\
\hline
CR & 0.786 & 0.797 & 0.774 & 0.791 \\
MR & 0.749 & 0.761 & 0.753 & 0.757 \\
\hline
\end{tabular}}
\end{table}

The performance of the proposed supervised fine tuning framework is detailed in Table \ref{perf}. All word embeddings are significantly enhanced after the supervised fine tuning. The performance of HLBL50 embedding is similar to the SENNA50 and Skip50 embeddings, although the initial performance of HLBL50 is not satisfactory. The performance of the best word embedding Glove300 in this experiment is also significantly improved in all datasets. These remarkable improvements may demonstrate that the supervised fine tuning framework may transfer the complementary knowledge from the weak embeddings and lexical semantic resources into the strong embeddings.

In the perspective of evaluation tasks, the result data supports the conclusion that this supervised fine tuning framework is effective for all the involved tasks. Due to the training label is strongly related to the similarity prediction tasks, the performance for all 5 datasets are significantly improved. The improvement for the analogical reasoning task is not as large as the similarity prediction task, but also remarkable. An interesting finding is that the performance of sentence completion task forms two groups, the word embeddings in high dimension obtain more significant improvement than the low dimensional word embeddings. The authors believe that this is caused by both the learning capability difference and overfitting.

Table \ref{sa} shows the result for sentiment analysis task. No matter the word vectors are directly used as the features or as an initialization of feature parameters, the fine tuned GloVe word embedding outperform the original one for both datasets. The may reveal that the fine tuned word embeddings have better separation capability than the original embeddings for sentiment analysis tasks.
\begin{figure}[!htb]
\begin{center}
\includegraphics[width=\columnwidth]{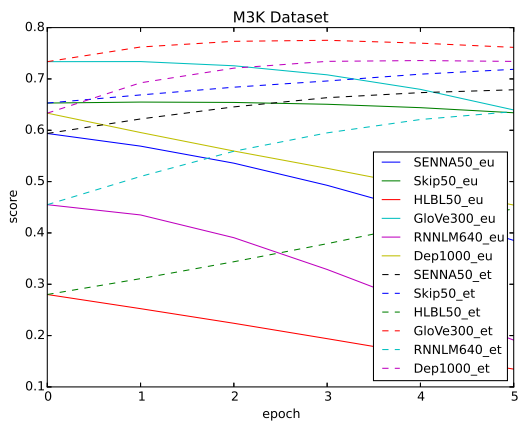}
\caption{Comparison of standard SGD and inverse error weighted SGD on M3K similarity prediction dataset..}
\label{simi}
\end{center}
\end{figure}

\begin{figure}[!htb]
\includegraphics[width=\columnwidth]{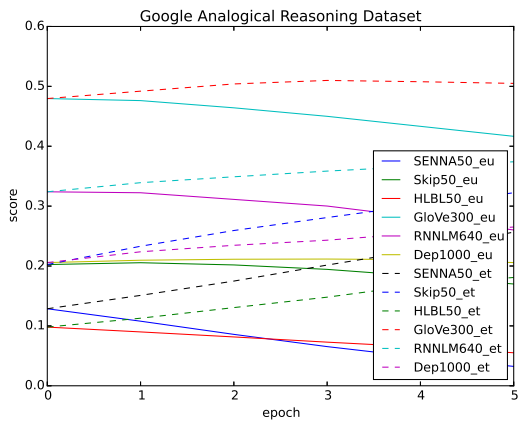}
\caption{Comparison of standard SGD and inverse error weighted SGD on google analogical reasoning dataset.}
\label{google}
\end{figure}

\begin{figure}[!htb]
\includegraphics[width=\columnwidth]{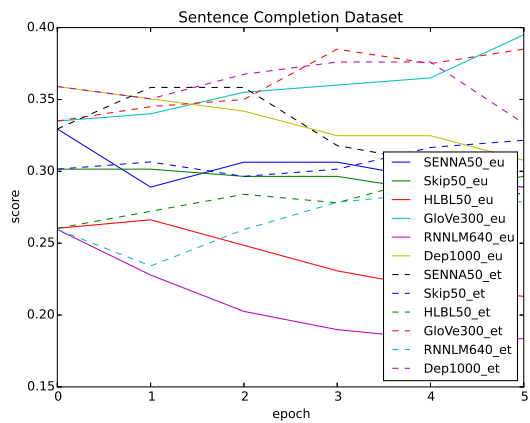}
\caption{Comparison of standard SGD and inverse error weighted SGD on sentence completion dataset.}
\label{sent}
\end{figure}
\subsubsection{Comparison of Standard SGD and Inverse Error Weighted SGD}

In the second experiment, the standard mini-batch SGD and the inverse error weighted mini-batch SGD are compared. The learning curve of both approachers for all embeddings on three tasks are shown in Figure \ref{simi}, \ref{google} and \ref{sent}, respectively.

It is shown that nearly all the solid lines representing standard mini-batch SGD in these three figures have a trend of decreasing, which means the standard mini-batch SGD does not help, but decreases the performance. In contrast, the performance of inverse error weighted approacher represented by dash lines have a general trend of increasing. This experiment demonstrates the phenomenon described above that the data with large error is too difficult to learn and may disrupt the original geometric similarity. With the obvious large performance gap in the comparison, the conclusion may be obtained that the inverse error weighted mini-batch SGD optimization algorithm is suitable for supervised fine tuning learning of word representation.
\subsubsection{Effect of Training Data Size}
\begin{table*}[!htb]
\centering
\caption{Effect of Data Size.}
\label{datasize}
\resizebox{\textwidth}{!}{%
\begin{tabular}{|c|c |c|c| c|c| c|c |c|c |c|}
\hline
Embedding & \multicolumn{5}{|c|}{GloVe300} & \multicolumn{5}{|c|}{Skip50} \\
\hline
Data Size & 0 & 50 &100 &150 &200 & 0 &50 &100 &150 &200 \\
\hline
WordSim353& 0.55 &0.58 &0.58 &0.61 &\textbf{0.62}& 0.53 &0.50&0.51 &0.51&\textbf{0.55} \\
Mturk771 &0.63&0.70 & 0.69&0.70 &0.69&0.52 &0.63&0.62 &0.60& 0.60   \\
RG65 & 0.74&\textbf{0.80} & 0.77& 0.70 &0.77& 0.56 &\textbf{0.67}& 0.64 & 0.63 &0.60 \\
YP130 & 0.55&0.62 &\textbf{0.63}& 0.59 &0.59& 0.35 &\textbf{0.56}& 0.52 &0.49& 0.45 \\
M3k & 0.73&0.76 &0.75& 0.78 & 0.77& 0.65 & 0.73& 0.73 & 0.73& 0.73 \\
Google\_AR &0.479&0.472 &0.479&0.494&\textbf{0.508}& 0.202 &0.295 & 0.33 &0.358 & \textbf{0.360}\\
MS\_AR & 0.665&0.67 &0.69&0.68 & \textbf{0.685}& 0.286  &0.395& 0.440 &0.446& \textbf{0.451}\\
MS\_SC & 0.335&0.42 & 0.39& 0.39 &0.375&0.30 &0.31& 0.30 &0.30& 0.30\\
\hline
\end{tabular}}
\end{table*}

Table \ref{datasize} show the effect of different size of training data. Obviously, the performance of the analogical reasoning tasks is improved slightly with the increase of training data size. However, the performance for similarity prediction tasks reveals two diverse paths, the small datasets like RG65 and YP130 prefer small size of training data, and the performance of large datasets like M3K and Mturk771 is not affected by the size of training data.

Generally, the increase of training data size is only slightly helpful to analogical reasoning tasks, so it empirically proves the hypothesis that employing only top related data is enough to support this supervised fine tuning.

\section{Conclusion}
A supervised fine tuning framework is proposed to boost the unsupervised word representation learning algorithms. The proposed automatic label generation approach and the inverse error weighted mini-batch SGD optimization algorithm make it possible to provide additional supervised fine tuning phase for all unsupervised word representation learning algorithms. Many experiments involving 10 datasets and 6 well trained embeddings empirically demonstrated that the framework is very effective for improving the quality of word representation.

\bibliography{ref}
\bibliographystyle{acl2012}

\end{document}